# Explainable Artificial Intelligence for Process Mining: A General Overview and Application of a Novel Local Explanation Approach for Predictive Process Monitoring

Nijat Mehdiyev[1,2], Peter Fettke[1,2]

[1]German Research Center for Artificial Intelligence (DFKI), Saarbrücken, Germany
[2] Saarland University, Saarbrücken, Germany
nijat.mehdiyev@dfki.de, peter.fettke@dfki.de

**Abstract.** The contemporary process-aware information systems possess the capabilities to record the activities generated during the process execution. To leverage these process specific fine-granular data, process mining has recently emerged as a promising research discipline. As an important branch of process mining, predictive business process management, pursues the objective to generate forward-looking, predictive insights to shape business processes. In this study, we propose a conceptual framework sought to establish and promote understanding of decision-making environment, underlying business processes and nature of the user characteristics for developing explainable business process prediction solutions. Consequently, with regard to the theoretical and practical implications of the framework, this study proposes a novel local post-hoc explanation approach for a deep learning classifier that is expected to facilitate the domain experts in justifying the model decisions. In contrary to alternative popular perturbation-based local explanation approaches, this study defines the local regions from the validation dataset by using the intermediate latent space representations learned by the deep neural networks. To validate the applicability of the proposed explanation method, the real-life process log data delivered by the Volvo IT Belgium's incident management system are used. The adopted deep learning classifier achieves a good performance with the Area Under the ROC Curve of 0.94. The generated local explanations are also visualized and presented with relevant evaluation measures that are expected to increase the users' trust in the black-box-model.

# 1    Introduction

In order to attain the competitive edge, enterprises are called upon to consistently and sustainably improve their abilities to accelerate time-to-market, to ensure the quality of products and services and to increase the scale of their business operations [1]. For these purposes, securing a robust integration of various business processes as well as maintaining data consistency upon such incorporation is the primary driver of success. As more and more processes are being equipped and instrumented with a wide variety of information systems and corresponding data sources, the ability to incorporate rapidly increasing volume of heterogeneous data into decision-making processes has become an indispensable prerequisite for managing processes efficiently.

The prevalence of process-aware information systems facilitates the capturing of digital footprints which are generated throughout the various stages of business processes. Process mining has recently emerged as an established research discipline aiming at delivering useful insights by using fine granular event log data delivered by enterprise information systems; the new discipline lies at the intersection of artificial intelligence particularly data mining and business process management [2]. Although the initial approaches of process mining have been primarily concerned with process analyses of a descriptive and diagnostic nature such as discovery of the business processes or bottleneck analysis, a substantial shift towards prospective analyses has recently been observed. The process owners exhibit in particular a considerable interest in the opportunities generated by predictive business process management, one of the rapidly evolving process mining branches, because such intelligence enables deviations from desired process execution to be detected proactively and it establishes a basis for defining prompt intervention measures.

The adoption of advanced machine learning methods facilitates to deliver robust, consistent, and precise predictions for examined targets of interest by learning the complex relationships among process sequences, process specific features, and business case related characteristics [3]. Nevertheless, due to their black-box nature these techniques suffer notably in delivering appropriate explanations about their outcomes, internal inference process and recommended courses of actions. Consequently, their utilization potentials for predictive business process management are substantially impaired, since the verification of validity and reliability to the models cannot be accomplished and the justification of individual model judgements cannot be ascertained which leads to lack of trust and reliance in the machine learning models. The recent approaches from the explainable artificial intelligence (XAI) research domain pursue the objective of tackling these issues by facilitating a healthy collaboration between the human users and artificial intelligent systems. Generating relevant explanations tailored to the mental models, technical and business requirements along with preferences of the decision makers is assumed to alleviate the barriers that the data-driven business process intelligence can be operationalized.

In this manuscript, we aim to examine the applicability and implications of the explainable artificial intelligence for process mining particularly for predictive business process management to lay foundations for prescriptive decision analytics. The main contributions of this study are multi-faceted:

As a first contribution, this study proposes a conceptual framework which is sought to guide the researchers and practitioners in developing explanation approaches for predictive process monitoring solutions. Being used as guideline it is expected to enable the designers and developers to identify the objectives of the explanations, the target audience, the suitability of identified methods for the given decision-making environment, and the interdependencies among all these elements. Approaching to the development process of explanation systems in such a systematic manner is crucial as previous research on the explanations in the expert systems suggests that the specifications regarding the explanation have to be considered during the early design of the intelligent systems, otherwise the quality of the generated explanations are likely considerably deteriorated [4]. E.g. the generated explanations may not meet the end users' requirements by being too complex, time-consuming, not necessarily relevant and imposing high cognitive loads on the users which reduce their usability, and even lead to failure in adoption of the underlying artificial advice givers [5]. For this purpose, this study attempts to provide an initial holistic conceptual framework for examining the decision-making environment and assessing the properties of the explanation situation before developing or adopting a specific explanation method for predictive process monitoring.

As a second contribution, this study instantiates the proposed framework for analyzing the implications of the examined use-case scenario and the characteristics of the underlying process data-driven predictive analytics on developing an appropriate explanation method. After defining the relevant explanation requirements, a novel local post-hoc explanation approach is proposed to make the outcomes of the adopted deep learning based predictive process monitoring method understandable, interpretable, and justifiable. One of the novelties in our proposed approach lies in the identification process of the local regions since the extracted intermediate representations of the applied deep neural networks are used to identify the clusters from the validation data. The local surrogate decision trees are then applied to generate relevant explanations. By following this approach, the proposed explanation method pursues the goal to overcome the shortcomings of popular perturbation based local explanation approach while using the model information as well. Furthermore, this study incorporates the findings from the cognitive sciences domain to make the generated explanations more reliable and plausible.

The remainder of the manuscript is organized as follows. Section 2 provides an overview of background and related work to process mining particularly by focusing on predictive business process management methods and application of the explainable artificial intelligence especially for deep learning approaches. Section 3 introduces the framework which can be used to carry out the explainable process prediction projects and provides a brief overview of chosen use-cases. In the Section 4, we present the proposed post-hoc local explanation approach for business process prediction scenario and discuss its stages in detail. The Section 5 introduces the examined use case from the incident management domain, discusses the details of the evaluation measures for the black-box, clustering and local surrogate models and finally highlights the obtained results and explanations. Section 6 discusses the scientific and practical implications, limitations, and future work. Finally, Section 7 concludes the study with a summary.

## 2   Background and Related Work

### 2.1   Process Mining

The adoption and deployment of advanced analytics approaches deliver sustained flexibility, precision, and efficiency in business processes, and thereby generates substantial added value along the entire value chain. Nevertheless, despite the fact that data-driven analytics has already been established as a capable and effective instrument and has found successful applications in diverse disciplines, they rarely embrace a business process perspective throughout the information processing and inferencing cycle. The process mining techniques have recently proven to provide a resilient mechanism to address this challenge by bridging the gap between data and process science. Various process-aware information systems such as Customer Relationship Management (CRM), Enterprise Resource Planning (ERP), Workflow Management Systems (WMS), Manufacturing Execution Systems (MES), Case and Incident Management Systems have the ability to log the process activities generated during the process execution. The delivered event log contains the information about the process traces representing the execution of process instances which are also referred as to cases. A trace consists of sequence of process activities which are described by their names, by timestamp and eventually by other information such as responsible people or units if available. Typical process data is similar to sequence data but due to branching and synchronization points, it get much more complex [6].

Various process mining methods such as process discovery, conformance checking, process enhancement, predictive business process management etc. use the event log data to generate important insights into business processes. The main purpose of the *process discovery* is the automatic data-driven construction of business process models from the event logs [7]. Heuristics mining, genetic process mining, region-based mining, alpha, alpha+, inductive mining etc. are different algorithms that can be used to synthesize process models from log data [8]. The *conformance checking* pursues the objective to examine the real process behavior by comparing the process models with the event log of these processes. Various alignment strategies and fitness measures can be used to compare the observed processes and process models that are hand-made or discovered from the event data [9]. The main idea behind *process enhancement* is extending the a-priori process models by analyzing the event logs. Generating process improvement strategies by analyzing the bottlenecks, service levels, throughput time etc. is a typical example for process enhancement [7].

### 2.2   Predictive Business Process Management

Predictive business process management also referred to as predictive process monitoring or business process prediction is another branch of process mining that aims at predicting the pre-defined targets of interest by using the activities from the running process traces [10]. The primary underlying concept of the predictive business process management is the anticipation of the user's defined target of interest with the usage of the process activities from the running cases. Several studies have been proposed to address the predictive process monitoring focusing on classification problems such as:

- **next event prediction** [6, 11–18]
- **business process outcome prediction** [19–22]
- **prediction of service level agreement violations** [15, 23]

There are also various studies that handle various regression problems in the business process prediction domain:

- **remaining time prediction** [24–27]
- **prediction of activity delays** [28]
- **risk prediction** [29, 30]
- **cost prediction** [31]

These studies use different machine learning approaches such as decision trees [21], support vector machines [16], Markov models [12], evolutionary algorithms [14] among others.

### 2.3 Deep Learning for Predictive BPM and XAI

In the light of reported predictive process analytics experiments and corresponding results, it is conceivable to state that deep learning approaches provide superior results to alternative machine learning approaches. Various deep learning architectures such as

- **deep feedforward neural networks** [3, 11, 32, 33]**,**
- **convolutional neural networks (CNN)** [34–38]**,**
- **long-short term memory networks (LSTM)** [6, 17, 33, 35, 39–41]**, and**
- **generative adversarial nets** [42]

have already been successfully implemented for different classification and regression tasks for predictive process monitoring problems.

Although these advanced models provide more precise results compared to conventional white-box approaches, their operationalization in the process mining applications suffers from their black-box nature. Lack of understandability and trust into these non-transparent models results in an enlarging gap between advanced scientific studies and their low adoption in practice. Recently there has been considerable interest in making the deep learning models understandable by using different explanation techniques in other research domains. A systematic analysis by [43] provides a comprehensive overview of explanation methods for deep neural networks by categorizing them as techniques explaining how the data are processed, approaches explaining the representation of the data and explanation producing systems. In another study by [44], a brief overview of various local explanation approaches for deep learning methods such as Input-Output Gradient [45], Integrated Gradients [46], Guided Backpropagation [47], Guided Grad-CAM [48], SmoothGrad [49] are provided and the sensitivity of these approaches is examined. A further comparative analysis of RISE, Grad-CAM and LIME approaches for making deep learning algorithms explainable can be found in [50]. An extensive literature analysis carried out by [51] focuses especially on visualization approaches by categorizing the identified studies into node-link diagrams for network architecture, dimensionality reduction and scatter plots, line charts for temporal metrics, instance-based analysis and exploration, interactive experimentation, algorithms for attribution and feature visualization classes.

Although generating explanations for deep learning and other black-box approaches have already attracted an increased attention in other domains, there are just a few approaches for explainable artificial intelligence in the process mining domain. The study by [52] proposed to generate causal explanations for deep learning-based process outcome predictions by using a global post-hoc explanation approach, partial dependence plots (PDP). The explanations are generated for process owners who would like to define long-term strategies like digital nudging by learning from the causal relationship among transactional process variables. The study by [53] introduced and illustrated the necessity of explainable artificial intelligence in a manufacturing process analytics use case. The authors of the article [54] applied the permutation based local explanation approach, LIME, to generate explanations for business process predictions. In a brief position paper the relevance of explainable artificial intelligence for business process management is highlighted as well [55]. However there is still much work to be done for explainable predictive process monitoring and thus to fill this gap, we propose first a framework which is expected to facilitate the guidance for choosing the appropriate explanation approaches and illustrate the applicability of the proposed novel local post-hoc explanation approach in a real-world use case.

## 3    A Framework for Explainable Process Predictions

Notwithstanding the fact that explainable artificial intelligence has just recently emerged as one of the major research disciplines in artificial intelligence research domain, the need to generate the explanations of intelligent systems is not a new phenomenon. Despite the fact that considerable amounts of studies have already been carried out over the last three decades, extensive investigations reveal that most attempts at explaining the black-box systems have not been entirely satisfactory due to their insufficient functionality to meet the users' requirements of understandability. Explainability is a broad and vague notion, as it encompasses a wide range of dimensions and objectives. To a considerable extent, their perceived adequacy and appropriateness rely on the prevailing circumstances of the decision-making environment and the nature of the user characteristics. Due to the inherent socio-cognitive nature of the explanation process, it is necessary to ensure an intensive interaction of decision makers with the underlying intelligent systems and produced explanations [56]. In this context it is essential to facilitate an adequate in-depth guidance that enables the human audience to clearly comprehend the nature and the causes of the examined phenomenon.

To cope with such a multi-faceted challenge, it is essential to systematically approach the explanation generation process by considering several relevant perspectives in a comprehensive way. Recent studies imply that the ongoing research and development initiatives pursued in the XAI area to this day still overlook numerous factors [57]. Apart from the researcher's intuition concerning the soundness and reasonability of an explanation, in many cases process-specific, human-related, organizational, and economic considerations as well as the inherent properties of explanations and problem settings are disregarded. Therefore, there is a need for a holistic framework that can be used as a guidance for design and development of explanatory systems. To overcome these challenges, we propose a conceptual framework by analyzing, combining and adapting the propositions from the explainable artificial intelligence research to process mining domain (see Figure 1). Below we provide a thorough discussion of the framework elements, *subject, objectives, techniques, generation time, outcomes,* and present the close links and implications of them to each other by illustrating examples.

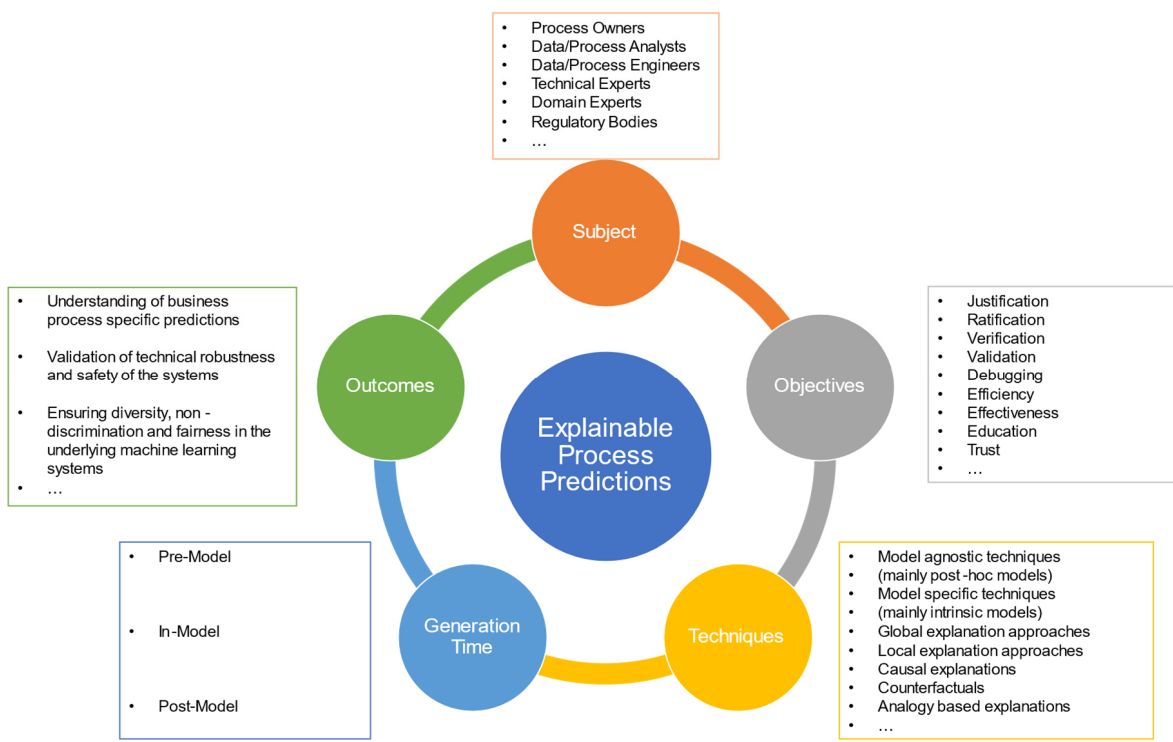

**Fig. 1.** A Conceptual Framework for Explainable Process Prediction

**Subject**: The nature of the explanations is significantly influenced by the group of the people having different backgrounds, motives, beliefs, and expectations who may use them for different decision-making situations. A series of recent research has merely concentrated on the implications of explanation recipients for choosing the proper explanation techniques. The study by [58] which is partially built on various suggestions from the literature [56, 59] defined six different types of the subjects. These subject types include the system *creators* developing the machine learning systems, system *operators, executors* making decisions by using the system outputs, *decision subjects* for whom the decisions are made, the *data subjects* whose data are used for training the models and *examiners* such as auditors. By conducting a more global analysis, the study by [60] investigated the main stakeholders for explainable artificial intelligence. According to their analysis *developers, theorists, ethicists and end-users* are the main subject groups which use or contribute to the process of generating explanation for intelligent systems. Similar to the studies outlined above, the study by [51] organizes the subjects into non-mutually exclusive groups and defines three main categories: *model developers and builders, model users and non-experts.*

To carry out process mining projects especially the ones with the predictive analytics features, various stakeholders have to be involved. Data/process engineers are the main developers and knowledge engineers who are responsible for creating and productionizing the machine learning based predictive process analytics systems. The data/process analysts are the executors or the model users who use the underlying systems to formulate and verify the hypotheses, make decisions and recommendations. The process owners are mainly non-experts in terms of the machine learning but with their specific knowledge about the processes they define the scope of the project and identify the technical and business success criteria. The domain experts who have deep expertise in the examined business processes may provide valuable feedback for both process engineers and analysts when carrying out their specific tasks. The technical experts are mainly responsible for a secure extraction of the process specific data from the process aware information systems and providing an initial understanding of the relevant data fields. Supervisory boards and regulatory bodies have also their specific interests on the structure and content of process predictions especially related to compliance and fairness issues.

**Objectives**: The objectives of the machine learning explainability which are mainly driven by the requirements of various subject groups are multifaceted and have significant implications for the success of the analytics project. A systematic analysis of the failed expert systems by [61] has revealed that the main reason for the failure was the inability of these systems to address the users objectives in demanding explanations. The findings by [62] suggest that the experience levels of the experts imply various considerations for generated explanation. The novice users prefer justification based terminological explanations by avoiding explanation of multi-level reasoning traces which may lead to high cognitive loads. According to [63], *verification* of the system capabilities and reasoning process, investigating the knowledge of the system which was coined as *duplication*, and *ratification* which aims to increase the trust of the end users in the underlying intelligent system are the three main objectives. Recent study by [64] which conducted an extensive analysis of the explainable AI tools and techniques have identified multiple other objectives. According to their findings, explanation objectives can be defined as explaining how the system works (*transparency*), helping the domain experts to make reasonable decisions (*effectiveness*), convincing the users to invest in the system (*persuasiveness*), increasing the ease of the use (*satisfaction*), allowing the decision makers to learn from the system (*education*), helping the users to make fast decisions (*efficiency*), enabling the decision makers to detect the defects in the system (*debugging*) and allowing the user to communicate with the system (*scrutability*).

**Techniques**: Over time, an extensive literature has developed on explainable artificial intelligence methods. Various systematic survey articles can be found in the studies by [65–69]. The explanation approaches can be categorized in accordance with various criteria. In terms of model relation, the explanation approaches can be model-specific implying that they are limited to the underlying black-box model and can be model-agnostic generating the relevant explanations independent of the used machine learning model. The model specific approaches by nature generate intrinsic explanations which can be used for verification purposes whereas the model-agnostic approaches have mainly post-hoc explanation character and facilitate the users in justifying the models and their outcomes. An alternative way to classify these methods is to differentiate them according to their scope. Global interpretability approaches *PDP*, *global surrogate models* [65, 70], *SHAP dependent plots* [71], *Accumulated Local Effects (ALE) Plots* [72] pursue the objective of providing explainability of the model for whole data observations whereas the local explanation approaches such as *Individual Conditional Expectation (ICE) Plots* [73], *Shapley Additive Explanations (SHAP)* [74], *LIME* [75] enable to examine the instances individually or defined local regions. Furthermore, depending on the objectives of the explanations and user preferences visual and textual, analogy/case-based explanations, causal explanations, counterfactuals can be generated.

**Generation Time**: The explanations can be generated before building the models which is referred as to pre-model, during the model building which is called in-model and after building the underlying machine learning model (post-model) [69]. The pre-model explanation approach is strongly related to exploratory data analysis and includes mainly visualization and graphical representation approaches. In-model methods which are mainly model specific intrinsic approaches, attempt to generate explanations during the training phase by testing e.g. constraint options. Finally, the post-model approaches are adopted once a strong predictive model is obtained and aim to generate explanation without examining the reasoning trace or inference mechanism of the black-box model.

**Outcomes**: Finally, the generated explanations can be used to get understanding in terms of different outcomes. Process analysts, domain experts and process owners may use the generated explanations to understand various business process specific analyses. By using causal explanations generated for next event predictions, the users can identify the limitations related to the process and user behavior and consequently define corresponding intervention measures. Justifying the process outcome predictions may facilitate to define the course-of-action for enhancing the business processes. Making resource specific prediction comprehensible would enable to generate various resource allocation strategies more efficiently. Interpretability may allow to examine the reasons for the deviations in process performance indicators (time delays, increasing costs). Furthermore, in order to realize the concept of the trustworthy artificial intelligence, it is very important to validate the technical robustness of the machine learning systems, to verify that the underlying model fulfills the required privacy and safety requirements and finally to ensure the fairness, non-discrimination and diversity.

**Potential Use-Case for Explainable Predictive Process Monitoring:** It is very essential to emphasize that these elements should not be considered in an isolated manner since they have direct and contextually strong relationships. Figure 2 provides an overview of chosen use-cases for predictive business process management which illustrates the links among these various dimensions of the analytics situation.

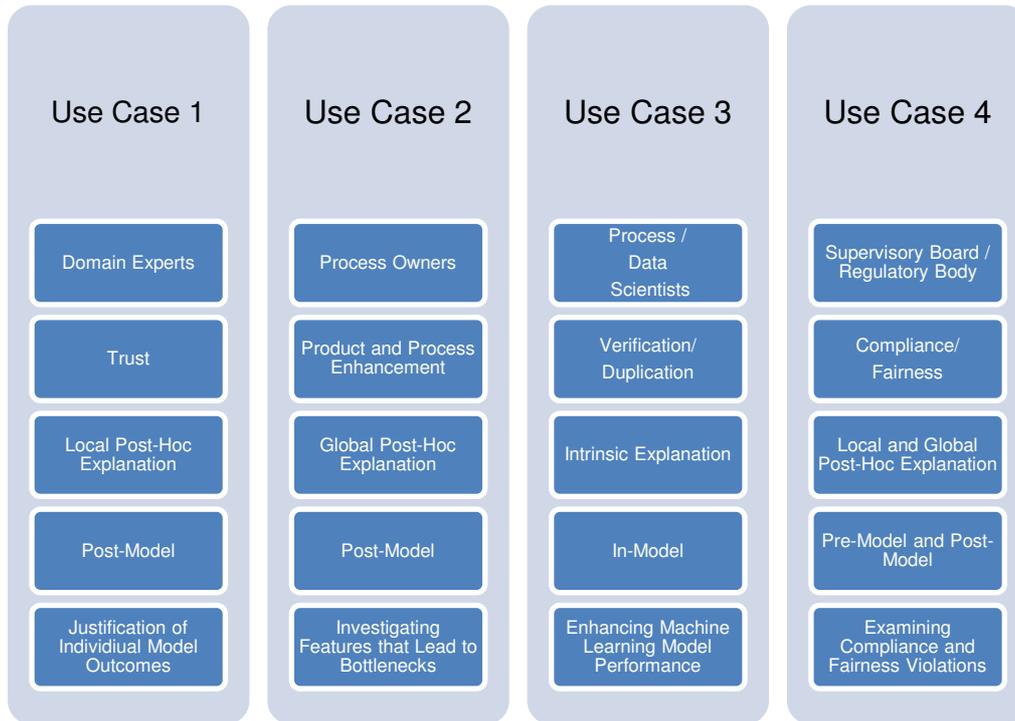

**Fig. 2.** Chosen Use-Cases for Explainable Process Predictions

In the first scenario, the target audience for explanations is defined as domain experts with limited machine learning background who aim to use the explanations to justify the model outcomes for individual observations. These users have limited interest and capability to understand the inner working mechanisms of the adopted black box models. Therefore, it is reasonable to develop relevant local post-hoc explanation approaches for them. In the second scenario, the process owners are provided with explanations that enable them to make more strategic decisions rather than focusing on each decision instance separately. Thus, it makes more sense to generate global post-explanation solutions that facilitate the process owners to understand relationships among features and model outcomes for the whole dataset. Once the relevant predictive insights are generated, they can decide how to enhance processes or improve the products and services. According to the third scenario, the process and data scientists aim to improve accuracy of the applied black-box prediction approach. Since these stakeholders are mainly interested in the reasoning trace of algorithms, the generated intrinsic explanations create more added value. Finally, the supervisory board or regulatory bodies are interested to understand whether using the adopted data-driven decision making violates the compliance or fairness criteria. For this purpose, it is reasonable first to provide the explanations for all model outcomes and then allow them to examine the specific individual model decisions. It is also essential to note that these are just some of the chosen illustrative use cases and the list can be easily extended by various stakeholders or various objectives, tools, outcomes etc. for the target audience discussed above. Furthermore, in many cases the combination of various explanation approaches provides more comprehensible and precise tool by facilitating the users to examine the process predictions from different perspectives. However, for this goal it is crucial to ensure smooth transitions among various explanation types since misleading explanations are worse than no explanation at all [4].

## 4    A Novel Local Post-Hoc Explanation Method

In our study, we propose a novel explainable process prediction approach after identifying key requirements and elements by using the conceptual framework introduced in the Section 3. The ultimate objective of the proposed method is defined as developing an explainable deep learning model for identifying the deviations from the desired business processes by using a real-world use case (see Section 5.1.). For this purpose, it is important to design an explanation technique by targeting the end users such as domain experts or process owners rather than the knowledge engineers. On these grounds, it is reasonable for us to follow the ratification/justification objective of the machine learning interpretability by explaining why the provided results by the deployed deep learning approach are reliable (see Use Case 1 in Figure 2). Therefore, we propose a post-hoc local explanation which uses a novel technique for identifying the local regions. By using the generated local explanations, the domain experts are expected to understand the process behavior resulting in undesired outcomes and to justify the model decisions.

Figure 3 presents an overview of the proposed approach. After preparing the process event data by extracting the relevant process specific features and n-gram representations of the process transitions and defining the target of interest the deep learning model is trained to learn the complex relationships among them. The trained black-box model does not only deliver precise prediction outcomes but also extract useful representations by performing multiple layers of the non-linear transformations. The intermediate latent space representations obtained from the last hidden layer of the network are then used as input variables to the k-means clustering approach with the purpose to define the local regions. At the final stage, by using the original input variables and the prediction scores by deep learning approach, individual local surrogate decision trees are trained for the identified clusters to approximate the behavior of the main black-box classifier in that locality. The learned representations of this comprehensible surrogate model, namely the decision trees and consequently extracted rules are then provided to the domain experts who aim to justify the decision for individual instances. In the following subsections a discussion of the details for each used method is provided.

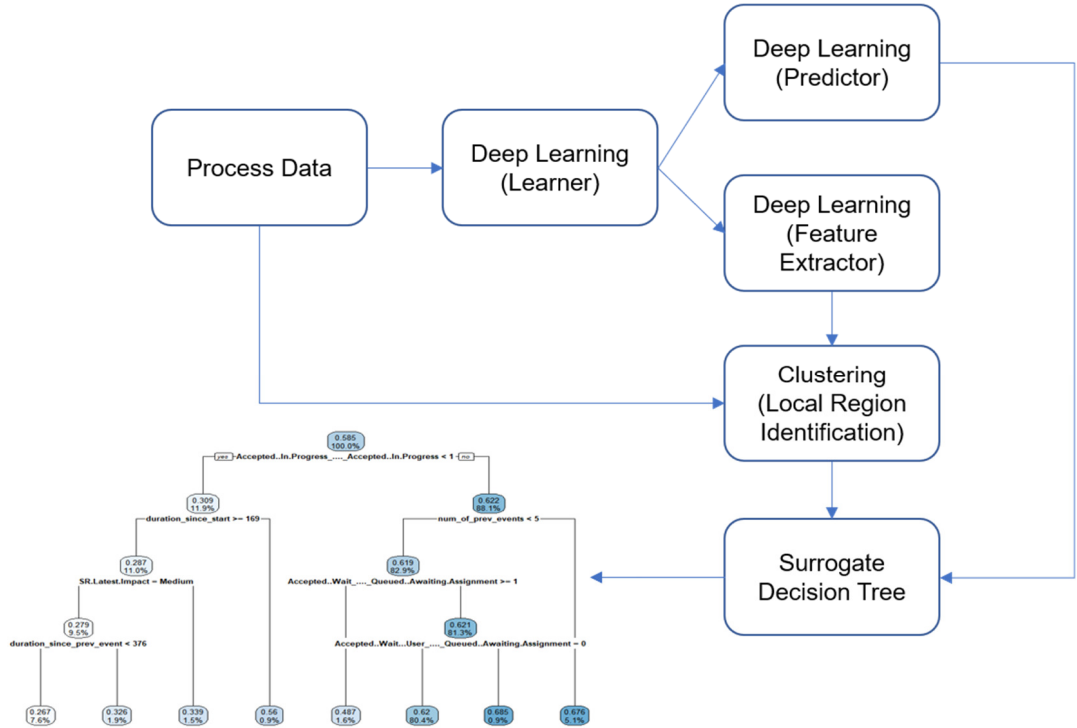

**Fig. 3.** Stages of the Proposed Local Post-Hoc Explanation Method

### 4.1 Binary Classification with Deep Learning

To generate plausible explanations, it is very important to ensure that a good predictive model is obtained. Considering the ability in addressing the discriminative tasks for different problem types from various domains and particularly for business process prediction problems, in our study we adopt a deep learning method as our black-box classifier. For the examined binary classification problem, we use the fully connected deep neural networks [76]. The stochastic gradient descent optimization (SGD) method was adopted to minimize the loss function and the specific lock-free parallelization scheme was used to avoid the issues related to this approach. Uniform adaptive option was chosen as an initialization scheme. To take the advantages of both learning rate annealing and momentum training for avoiding the slow convergence, the adaptive learning rate algorithm ADADELTA was used. Furthermore, we used the dropout technique (with rectifier) and early stopping metric as regularization approaches to prevent the overfitting in the underlying deep feedforward neural networks. The dropout ratio for the input layer was defined 0.1 whereas this value was set as 0.3 for hidden layers. The area under the ROC Curve (AUROC) was selected as stopping metric for which the relative stopping tolerance was defined at 0.01 and the stopping rounds as 10.

### 4.2 Local Region Identification by Using Neural Codes

Our proposed approach aims to generate the post-hoc explanations particularly local explanations. The purpose of local explanations is making the results of the black-box algorithms comprehensible by investigating the small local regions of the response functions since they are assumed to be linear and monotonic which lead to more precise explanations [77]. It was reported in literature that there are two main types of local explanation approaches [78, 79]. The model specific approaches such as saliency masks attempt to explain solely the neural networks whereas the model-agnostic approaches such as LIME, SHAP generate explanations independent of the underlying black-box models. Although these two perturbation-based approaches have recently gained an increased attention in the community and have found applications for various problem ranging from text classification to image recognition, they have been criticized for various issues. A recent study by [78] which investigated the robustness of the local explanation approaches revealed that the model-agnostic perturbation-based approach are prone to instability especially when used to explain non-linear models such as neural networks. Another study by [80] also concluded that in addition to high sensitivity to a change in an input feature, the challenges in capturing the relationships and dependencies between variables are the further issues that linear models suffer in the local approximation. A further limitation of these perturbation-based approaches is the random generation of the instances in the vicinity of the examined sample that doesn't consider the density of the black-box model in these neighbor instances [79]. In short, the literature pertaining to perturbation-based approaches strongly suggests that the local region should be identified and examined carefully to deliver stable and consistent explanations.

To address these issues outlined above, various alternative approaches have been proposed. The study by [79] used the genetic algorithms to identify the neighbor instances in the vicinity of the examined instance which is followed by fitting the local surrogate decision trees. The study by [81] proposed an approach that defines the local regions by fitting the surrogate model-based trees and generating the reason codes through linear models in the local regions that are defined by obtained leaf node paths. Similar to this approach, the K-LIME approach proposed by [82] attempts to identify the local regions by applying the k-means clustering approach on the training data instead of perturbed observation samples. For each cluster, a linear surrogate model is trained, and the obtained $R^2$ values are used to identify the optimal number of the clusters. One of the main critics to this approach is that the adopted clustering approach doesn't use the model information which leads to issues in preserving the underlying model structure [81].

In this study, we use a similar technique to the K-LIME however we follow a slightly different approach. The major difference lies in the identification procedure of clusters, where we do not use original feature space as input to the k-means clustering approach but non-linearly transformed feature space. Our method is inspired by the approach proposed by [83] which attempts to generate case based explanations to non-case based machine learning approaches. The main idea behind this technique is using the activation of hidden units of the learned neural networks during a discrimination task as distance metric to identify the similar cases or clusters from the model's point of view. The findings in [83] imply that using the Euclidean distance on the original input space to identify the locality as performed in the K-LIME approach can be useful to explain the training set to users. However, this approach fails to deliver plausible explanation regarding the trained model since it doesn't capture how the input variables are processed by the model. More specifically, in our approach in addition to its discrimination task, we use the deep neural networks as feature extractors by using the learned intermediate representations from the pre-last network layer as input to the unsupervised partitioning approach. Previous studies have shown that the using the latent space is a promising approach for neighborhood identification and the non-linear transformation approaches especially deliver promising results [84], [85], [86]. It is also worth to mention that even though the deep learning models such as deep autoencoders can be used to extract the latent space in an unsupervised fashion, we use the learned representation from the network that was trained to classify the business processes which is more relevant to generate the explanations as it preserves the used black-box model structure.

### 4.3 Surrogate Model

Once the local regions (clusters) are defined, at the last stage of our proposed approach we fit a local surrogate decision tree model. Referred also as to emulators, model distillation or model compression, the main idea of the surrogate models is approximating the behavior of the black box model accurately by using white-box models and providing explanations by introducing their interpretable representations. More specifically, in our case we use the deep learning model to generate predictions for the validation data. Following this for each cluster we train a decision tree by using the original variables from the validation set as input data and the prediction scores by deep learning model as output data. The learned decision paths and extracted rules provide cluster specific local explanations. To evaluate the approximation capability of the surrogate decision tree, we calculate the $R^2$ value for local surrogate decision trees in each cluster.

The position paper by [87] discussed the advantages and shortcomings of various white-box approaches including decision trees, classification rules, decision tables, nearest neighbors and Bayesian network classifiers by comparing their comprehensibility with use-based experiments. More specifically, the study by [88] compared the predictive performance of various comprehensible rule-based approaches such as RIPPER, PART, OneR, RIDOR etc. It is worth to mention that any of these alternative comprehensible, whit-box approaches can be adopted instead of decision trees as in the underlying study.

## 5    Experiment Setting

### 5.1    Use Case: Incident Management

The applicability of the proposed local explainable artificial approach is examined for an incident management use case by analyzing the real-world process log data delivered by Volvo IT Belgium's incident system called VINST [89]. The main purpose of the incident management processes is ensuring the quality of normal service operations within the Service Level Agreements (SLA) by mitigating potentially adversarial impacts and risks of incidents to the continuity of the service operations. Once the process owners verify that the intended level of the services has been attained, the incidents can be resolved. In case of suspected resurgence of such incidents, a problem management process should be undertaken with the aim of ascertaining their root causes and adopting the corresponding corrective and preventive procedures.

By using various information systems, the experts from different service lines perform their specific tasks to avoid the disruption to the business caused by incidents of various characteristics. One of such dimensions is the impact criticality of an incident, which is measured by the magnitude of the influence on the deterioration of the service levels and the number of concerned parties. The *major* impact incidents include the ones that disrupt the plant production, the business processes for designing, ordering and delivering the vehicles or negatively influence the cash flow and even the public image. The examples for *high* impact incidents are server incidents that result in customer loss or infrastructural incidents that may trigger production losses. In case the incidents only affect a limited proportion of the clientele and have no negative consequences on their service provision capabilities or only partially hamper internal IT services, they can be characterized as *medium* impact incidents. Finally, *low impact incidents* affect only a small number of customers and do not impede the functionality and performance if the necessary measures are followed. In accordance with their specific content and criticality of incidents three main service lines take various actions. The *first line* includes the service desk which is a first point of contact to provide efficient and expeditious solutions for customers to restore normal services and the expert desk where the application specific assistance is carried out. Incidents that cannot be handled by the first line staff are processed by the *second line*. The professionals on the *third line* are specialized in their product domain and may also operate as a support unit. Similarly, when incidents remain unresolved by the second line, the experts of third line will endeavor to resolve them.

The company pursues the objective and strategic vision that most of the incidents are to be addressed and successfully completed by the teams of the first service line. By attaining this objective, the efficiency of the work can be increased significantly. However, the process owners have noticed an improper usage of the "push-to-front" mechanism which is referred to the transfer of incidents to second and third service lines that could have been easily solved in the first line. Since such an overload of support activities that are not the main task of these experts in the second and third service lines, causes obstructions in the running of their core business activities it is essential to examine the dynamics of such "push-to-front" processes. To address this challenge, this study aims to build an efficient prediction approach and to extract corresponding explanations which may serve as as basis for generating proactive intervention measures. The proposed local explanation approach is expected to increase the transparency by extracting easily interpretable rules with low complexities and to improve the reliability to the underlying black box model by making each prediction decision explainable. The level of the trust in the black box machine learning system increases in case the extracted local decision trees and rules conform to domain knowledge and reasonable expectations [82].

### 5.2 Evaluation Measures

In this section, a brief overview of the evaluation procedures and measures for different stages of the proposed approach is given. After introducing the binary classification evaluation measures which are required to assess the performance of the deep neural network model, we discuss briefly how the fitness of the obtained clusters and the approximation capability of the adopted local surrogate decision trees can be measured. The predictive performance of the deep learning classifier for the examined classification problem can be assessed by calculating the number of correctly identified class instances (true positives), the number of correctly recognized instances that are not member of the class (true negatives), the number of the observations that are wrongly recognized as the class instances (false positives) and the number of the examples that were not identified as class members (false negatives). By using these measures, a confusion matrix also referred as to contingency table is constituted (see Table 1).

**Table 1**. Confusion Matrix

|  | Ground Truth Values | |
|---|---|---|
| **Predicted Values** | Positive | Negative |
| Positive | true positive *(tp)* | false positive *(fp)* |
| Negative | false negative *(fn)* | true negative *(tn)* |

To examine the predictive strength of the adopted classifier for the predictive process monitoring binary classification evaluation measures are computed (see Table 2).

**Table 2.** Binary Classification Evaluation Measures

| Measure | Formula | Measure | Formula |
|---|---|---|---|
| *Accuracy* | $\dfrac{tp + tn}{tp + tn + fp + tp}$ | *MCC* | $\dfrac{tp * tn - fp * fn}{\sqrt{(tp + fp) * (tp + fn) * (tn + fp)}}$ |
| *Precision* | $\dfrac{tp}{tp + fp}$ | *F1-Measure* | $\dfrac{2tp}{(2tp + fp + fn)}$ |
| *Recall* | $\dfrac{tp}{tp + fn}$ | *False Negative Rate* | $\dfrac{fn}{fn + tp}$ |
| *Specificity* | $\dfrac{tn}{tn + fp}$ | *False Positive Rate* | $\dfrac{fp}{tn + fp}$ |

Although these single-threshold classification evaluation measures provide useful insights into the model capabilities, a challenging problem arises in the identification of the correct threshold. For this purpose, it is important to define the cost function that should be optimized at the chosen threshold which in turn require a careful consideration of features and dimensions related to the decision-making mechanism. In the examined incident management use-case, the domain experts are interested in identifying the potential push-to-front activities more accurately but without sacrificing the general performance of the classifier. To address these requirements, we identify the threshold where the misclassification error is minimized across both classes equally. In addition to single threshold measures, the area under ROC Curve is also calculated and visualized. This threshold-free metric does not only give a more precise information about the model performance but also can guide for selecting the threshold more interactively. To assess the goodness and validity of the performed k-means clustering, a sum-of-squared based ratio is calculated. This ratio of sum-of-squares between clusters (SSBC) to sum-of-squares within cluster (SSWC) cluster is expected to measure total variance in the data that is explained by the performed clustering (see Table 3).

**Table 3.** Clustering Evaluation Measures

| Measure | Formula |
|---|---|
| Sum-of-squares between clusters (SSBC) | $\sum_{i=1}^{N} \|x_i - C_{p_i}\|^2$ |
| Sum-of-squares within cluster (SSWC) | $\sum_{i=1}^{M} n_i \|c_i - \bar{X}\|^2$ |

where $X = \{x_1, \ldots, x_N\}$ represents the data set with $N$ $D$-dimensional points, and $\bar{X} = \sum x_i / N$ is the center of the entire data set. The centroids of clusters are $C = \{c_1, \ldots, c_M\}$, where $c_i$ is the $i$-th cluster and $M$ is the number of clusters.

Finally, to estimate the goodness fit of the local surrogate model and to evaluate how good it can approximate the behavior of the global black box model in the identified cluster locality the $R^2$ measure is calculated (see Table 4).

**Table 4.** Surrogate Model Evaluation Measure

| Measure | Formula |
|---|---|
| $R^2$ measure | $1 - \dfrac{\sum_{i=1}^{n}(y_i - \hat{y}_i)^2}{\sum_{i=1}^{n}(y_i - \bar{y})^2}$ |

where $y_i$ represents the prediction score delivered by local surrogate model for the observation i, $\hat{y}_i$ denotes the black box prediction score for instance i and the $\bar{y}$ is the mean of black-box prediction scores.

### 5.3 Results

#### 5.3.1. Classification with Deep Learning

The examined process instances are randomly split in ratio 80:20 into training sets and validations sets. This section reports the evaluation results obtained on the validation set using the deep neural networks model. The Area under the ROC Curve (AUROC) is presented in the Figure 4. The applied black-box model achieves a remarkable performance with AUROC of 0.94.

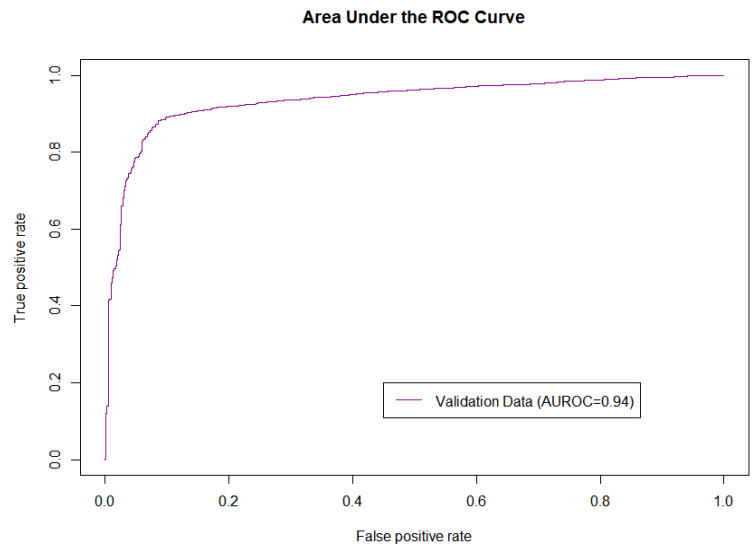

**Fig. 4.** Area under the ROC Curve

To carry out a more detailed investigation, we compute further several single threshold binary classification assessment measures in the predefined threshold of 0.9119 (see Table 5).

**Table 5.** Binary Classification Evaluation Results

| Measure | Formula | Measure | Formula |
|---|---|---|---|
| *Accuracy* | 0.8935 | *MCC* | 0.4704 |
| *Precision* | 0.9944 | *F1-Measure* | 0.9412 |
| *Recall* | 0.8934 | *False Negative Rate* | 0.1066 |
| *Specificity* | 0.8946 | *False Positive Rate* | 0.1054 |

The achieved results at this threshold, which equally minimizes misclassification across both classes, suggest that the push-to-front processes can be effectively and timely identified.

### 5.3.2. Local Explanations with Surrogate Decision Trees

The optimal number of the clusters is identified by maximizing the average local accuracy of the deep learning model in the clusters at the 0.9119 threshold. In the latest run (for which we also present the local results) the number of clusters was defined as 27. The ratio of SSBC/SSWC=0.915 indicates a good fit where the 91.5% is the measure of total variance that is explained by clustering.

After identifying the clusters, the decision trees are fit locally. In order to facilitate the users to justify the validity of generated explanations, they should be provided with various predictive analytics information including the prediction scores of the global deep learning model and surrogate decision trees for the examined instance, the $R^2$ value of local surrogate models, the ground truth label and obtained decision tree paths. With the following example we introduce the explanation for randomly chosen true negative prediction with relevant statistical properties. The Figure 5 presents the explanations for the chosen instance from the cluster 7 which was correctly classified having the following push-to-front activity.

| Supplementary Explanation Information | |
|---|---|
| Cluster number | 7 |
| Instance number | 18 |
| $R^2$ Value (for local surrogate model in the examined cluster) | 0.908 |
| Deep Learning Prediction | 0.290 |
| Surrogate Tree Prediction | 0.267 |
| Prediction | Push-to-Front |
| Ground Truth Label | Push-to-Front |

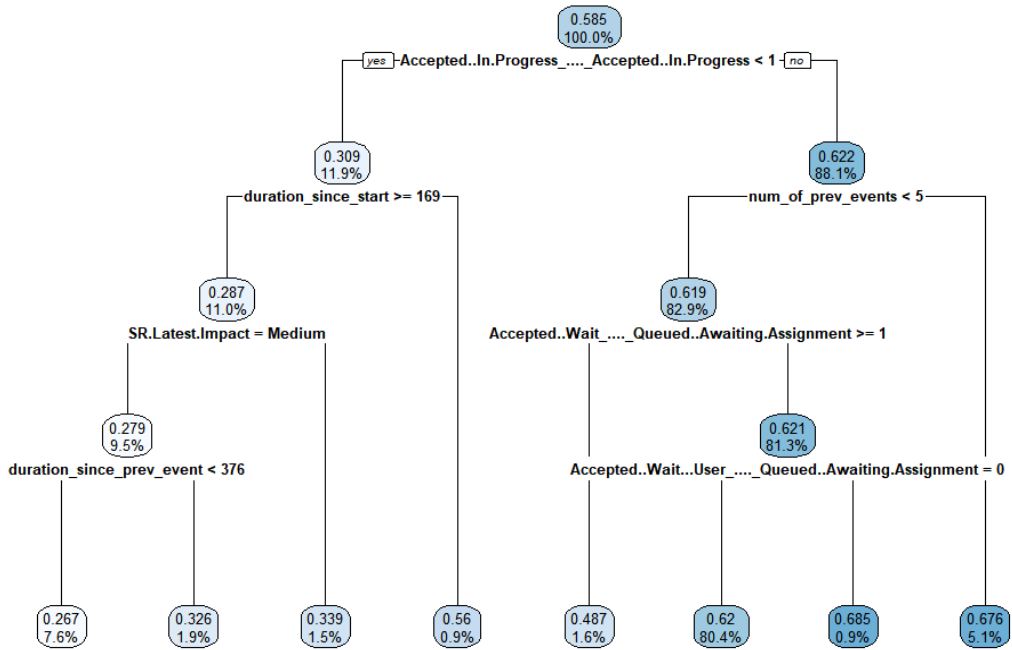

**Fig. 5.** Local Decision Tree

The applied deep learning approach confidentially classifies the instance by assigning a prediction score of 0.290 to the class "push-to-front" which is significantly below the pre-defined classification threshold (0.9119) that is supposed to optimize the selected cost function. The local surrogate decision tree model delivers a similar prediction score for the examined instance with 0.267 suggesting an acceptable approximation. Furthermore, the surrogate model has a $R^2$ value of 0.908 in the cluster to which the examined instance belongs to suggests that it is strongly capable to approximate the behavior of the applied deep learning in that local region. After verifying the validity of the global black box prediction and the suitability of the local surrogate model, the learned representations of the white-box-model, the obtained decision paths, and correspondingly extracted rules, are used for explanation purposes. The examined instance follows *Left-Left-Left-Left* path of the decision tree illustrated in the Figure 5. The extracted rule is as follows:

*IF the Accepted-In.Progress---Accepter-In.Progress is less than 1*
*AND duration since start is greater than 169 seconds*
*AND Impact is Medium*
*AND Duration since previous event is less than 376 seconds*
*THEN Prediction of Surrogate Model is 0.267*

Furthermore, it is very important to incorporate the findings of the cognitive sciences into explanation models and interfaces to eliminate the inherent cognitive biases. The study by [90] has empirically revealed that showing the confidence of the extracted rules leads to an increased plausibility of the explanation. Considering this suggestion, we present the confidence of each rule to the users. For the examined instance, the rule introduced above has a confidence of 0.60.

## 6  Discussion

This study aims to design an overarching framework which can be used to develop explainable process prediction solution by allowing to consider various aspects of the decision-making environment and the characteristics of the underlying business processes. For this purpose, an extensive analysis of the relevant studies from the explainable artificial intelligence domain was performed, the relevant findings were examined, combined, and customized for process mining applications. Although the proposed conceptual framework provides a solid fundament and preliminary basis for a more sophisticated understanding of the requirements for developing the explanation methods and interfaces, future studies could fruitfully explore this issue further by addressing other dimensions. The relevant activities in this context should be devoted e.g. to the defining the relevant evaluation mechanisms, procedures, measures, and methods. For this purpose, it is crucial first to define properties of the generated explanations such as fidelity, consistency, stability, comprehensibility, certainty, importance, novelty, stability, representativeness etc. by exploring the relevant literature [69, 91]. Furthermore, various taxonomies for interpretability evaluation such the one proposed by [92] which includes application grounded evaluation with real humans and real tasks, human-grounded evaluation with real humans, and simplified tasks and functionally-grounded evaluation with no humans, and proxy tasks should be positioned in the proposed framework adequately. Finally, providing an overview of measurement types in the evaluation user studies such as subjective perception questionnaires, explanation exposure delta, correct choices, learning score, domain specific metrics, choice of explanation, interest in alternatives, recommendation acceptance likelihood etc. would be beneficial for system developers to structure their evaluation procedures [64].

After presenting the details, we have instantiated the framework for developing an explainable process prediction approach for domain experts by using the data from the real-world incident management use case. For this purpose, a novel local post-hoc explanation approach was proposed, applied and evaluated. Compared to alternative perturbation-based local explanation techniques that suffer from high computational costs and instabilities, we proposed to identify the local region from the validation dataset by using the learned intermediate latent space representations delivered by the adopted deep feedforward neural networks during the underlying classification task. The idea of using the learned representations can be extended not only to alternative deep learning architectures but also can be easily adapted to alternative black box approaches such as random forest or gradient boosting approaches. For these ensemble tree-based approaches the leaf node assignments can be used as input to clustering approaches.

In the proposed approach, the decision tree was used as surrogate model to approximate the behavior of deep learning classifier. However, the alternative comprehensible methods such general linear models and various rule-based models can be implemented. Furthermore, it is worth to mention that the presentation of the learned rules or reason codes, which are intended to explain the model decisions, can imply various cognitive biases. For this reason, the findings from the cognitive sciences should be applied to develop appropriate debiasing techniques with the aim to increase the quality and plausibility of the explanations [90].

This study examined mainly the applicability of a local explanation approach in-depth which facilitates the domain experts to justify the individual model decisions. Such explanations are important to understand the process behavior and validate the judgment capabilities of the black-box approaches. The local explanation approach can also be used for alternative purposes such as understanding the reasons for discrimination once it is detected by using the relevant statistical measures. Furthermore, the global explanation approaches can be used which are mainly suitable for generating long term strategies for enhancing business processes.

## 7   Conclusion

In this study we proposed a novel local post-hoc explanation approach for predictive process monitoring problem to make the adopted deep learning method explainable, justifiable, and reliable for the domain experts. To ensure the suitability of this approach for the examined incident management use case, we also incorporated the relevant information from our proposed conceptual framework in this manuscript to the explanation generation process. Since the strong predictive performance of the adopted black-box model is a crucial prerequisite for the meaningfulness of the post-hoc explanations, we evaluated first our classier by using both threshold-free and single-threshold metrics. The area under ROC Curve of 0.94 and the accuracy of 0.89 at the threshold where the misclassification in both classes is minimized equally, suggest that a good performance model was obtained. After validating the performance of the model we introduced the explanations by presenting the local surrogate decision trees and showing relevant measures such as $R^2$ of the local model, the prediction scores from both black-box and surrogate model, the confidence of the rule etc. as well. By proposing a general framework for carrying out explainable process analytics and illustrating the applicability with a specific use case, we have attempted to emphasize the importance and relevance of explanation for process mining applications.

## Acknowledgment

This research was funded in part by the German Federal Ministry of Education and Research under grant number 01IS18021B (project MES4SME) and 01IS19082A (project KOSMOX).